# Long-Range Correlation Supervision for Land-Cover Classification from Remote Sensing Images

Dawen Yu, Shunping Ji, *Senior Member, IEEE*

*Abstract*—Long-range dependency modeling has been widely considered in modern deep learning based semantic segmentation methods, especially those designed for large-size remote sensing images, to compensate the intrinsic locality of standard convolutions. However, in previous studies, the long-range dependency, modeled with an attention mechanism or transformer model, has been based on unsupervised learning, instead of explicit supervision from the objective ground truth. In this paper, we propose a novel supervised long-range correlation method for land-cover classification, called the supervised long-range correlation network (SLCNet), which is shown to be superior to the currently used unsupervised strategies. In SLCNet, pixels sharing the same category are considered highly correlated and those having different categories are less relevant, which can be easily supervised by the category consistency information available in the ground truth semantic segmentation map. Under such supervision, the recalibrated features are more consistent for pixels of the same category and more discriminative for pixels of other categories, regardless of their proximity. To complement the detailed information lacking in the global long-range correlation, we introduce an auxiliary adaptive receptive field feature extraction module, parallel to the long-range correlation module in the encoder, to capture finely detailed feature representations for multi-size objects in multi-scale remote sensing images. In addition, we apply multi-scale side-output supervision and a hybrid loss function as local and global constraints to further boost the segmentation accuracy. Experiments were conducted on three public remote sensing datasets (the ISPRS Vaihingen dataset, the ISPRS Potsdam dataset, and the DeepGlobe dataset). Compared with the advanced segmentation methods from the computer vision, medicine, and remote sensing communities, the proposed SLCNet method achieved a state-of-the-art performance on all the datasets. The code will be made available at gpcv.whu.edu.cn/data.

## I. Introduction

Land-use and land-cover (LULC) classification assigns a semantic category for each pixel in a given remote sensing image, which supports geographic information system (GIS) mapping, environmental monitoring, urban planning, management, and change detection [1]-[4]. Currently, convolutional neural networks (CNNs) have become the mainstream in various image-based classification, object detection, and semantic segmentation tasks [5]-[7], including LULC classification. In particular, encoder-decoder style fully convolutional network (FCN) is the most popular framework in dense pixel-wise classification tasks [8], [9].

A remote sensing image covers a much broader scene than a medical or an indoor/outdoor image (see Fig. 1), which naturally leads to larger intraclass variance and may introduce smaller interclass variance, both of which are negative factors for accurate classification. Specifically, the differences between different pixels/entities of the same category can increase along with the increasing geospatial distance, which therefore requires a long-range information interaction mechanism to help a deep learning model be smarter. In previous studies, both attention mechanisms and transformer models have been introduced to model the long-range dependencies for the pixel-wise classification of remote sensing images [10]-[13]. However, all of these methods employ an implicit and unsupervised manner where they only calculate the self-correlation between pixels, and the correlation is not supervised. In fact, the long-range dependency modeling with explicit supervision, based on the available ground truth semantic map for a supervised learning task, may prove to be more efficient and has not been explored yet. This is the main focus of our work. Note that we discuss the long-range correlation modeling for the supervised learning in this paper, not including the unsupervised learning.

Remote sensing images not only have a large capacity but also contain objects of various sizes. Feature extraction at multi-scale levels has garnered significant attention in recent years due to its ability to provide reliable cues and capture intricate details. However, popular multi-scale information aggregation methods have primarily focused on the feature level [14]-[16] and operate on the single input image. While this is an efficient strategy, there is also a need to explore proper and efficient processing at the multi-scale image level to complement the feature-level operations. This is another problem that we address in our work.

In this paper, we propose a novel method called the supervised long-range correlation network (SLCNet) to model global dependencies for land-cover classification. SLCNet aims to capture long-range correlations in a supervised manner, thereby improving the accuracy of land-cover classification. In particular, SLCNet incorporates two essential modules. The first module is the long-range feature correlation supervision module (FCSM), which is responsible for establishing global

---

Manuscript was submitted on July 2, 2023. This work was supported by the National Natural Science Foundation of China (grant No. 42171430) and the State Key Program of the National Natural Science Foundation of China (grant No. 42030102) (Corresponding author: Shunping Ji.)

D. Yu and S. Ji are with the School of Remote Sensing and Information Engineering, Wuhan University, Wuhan 430079, China (e-mail: yudawen@whu.edu.cn; jishunping@whu.edu.cn).

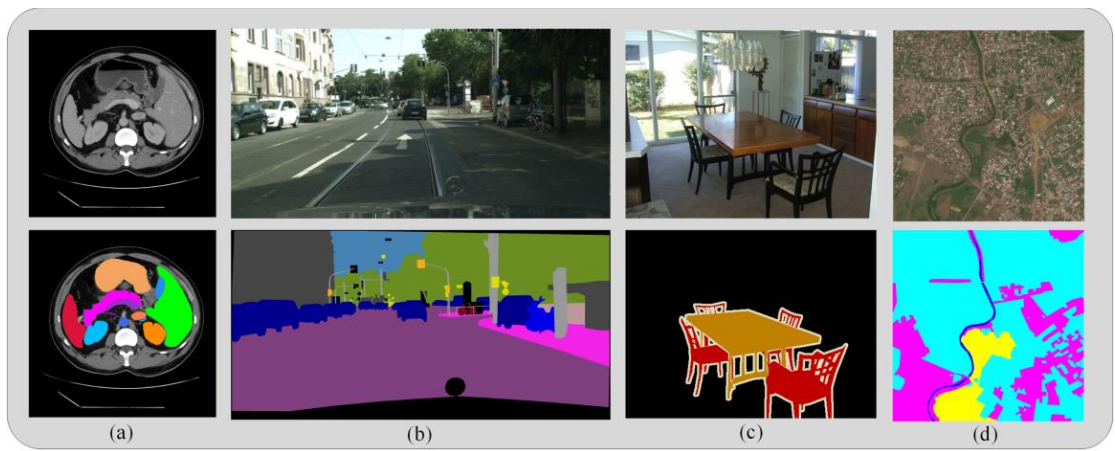

Fig. 1. The images and the corresponding semantic label maps: (a) medical image/label from the Synapse dataset[1]; (b) outdoor street image/label from the Cityscapes dataset[2]; (c) indoor dining room image/label from the VOCdevkit dataset[3]; (d) remote sensing image/label from the DeepGlobe land cover classification dataset[4].

long-range dependencies. The second module is the adaptive receptive field feature extraction module (ARFE), which operates at multiple image scales to capture local detailed features. By utilizing both modules, SLCNet ensures a complementary approach where FCSM emphasizes global information while ARFE captures local information effectively.

The main contributions of this paper are summarized as follows:

1) Supervised long-range dependency modeling for deep learning based semantic segmentation is investigated for the first time. Unlike the popular self-optimized attention-based long-range dependency modeling, our approach explicitly utilizes the ground truth semantic segmentation map to supervise the correlation scores between pixels based on their category labels. By doing so, we ensure that the recalibrated deep features are more consistent for pixels belonging to the same category, and more discriminative for pixels belonging to different categories, regardless of their proximity.

2) We propose a long-range feature correlation supervision module (FCSM) that implements the supervised global long-range dependency modeling. Additionally, we design an adaptive receptive field feature extraction module (ARFE) to capture size-adaptive detailed features from multi-scale images for accurately segmenting multi-size objects. FCSM and ARFE are parallelly integrated into a unified network called SLCNet. SLCNet has demonstrated exceptional performance in pixel-wise classification of remote sensing images.

## II. Related Works

In this section, we review previous studies for long-range dependency modeling and multi-scale feature extraction in remote sensing LULC classification. The limitations of existing approaches are also summarized.

### A. Long-Range Correlation Modeling for Remote Sensing Image Segmentation

Long-range information interaction modeling is very important for handling the challenge of the large intraclass variance and small interclass variance likely existing in a LULC classification problem. For the modern deep learning based classification methods, various approaches have been developed over the past few years to alleviate the intrinsic locality of standard convolution operations.

Some researchers have employed a CNN attention mechanism to model the long-range dependency. For example, Mou *et al.* [10] introduced a spatial relation module and channel relation module to learn the global relationships between two arbitrary spatial positions on the feature map. Niu *et al.* [11] designed three collaborative attention modules—the class augmented attention module, the class channel attention module, and the region shuffle attention module—to capture the global correlations, to produce more highly discriminative representations. Fu *et al.* [17] proposed a dual-attention network to capture contextual dependencies from both channel and spatial dimensions.

More recently, the transformer structures originating from natural language processing (NLP) have been modified and introduced into the computer vision field to model global dependencies [18], [19]. A transformer block usually consists of alternating layers of multi-head self-attention (MSA) modules and multiple-layer perceptrons (MLPs), and a LayerNorm layer is applied before each MSA and each MLP [18], [19]. The computation in the MSA layers in a transformer block is globally related, which makes it naturally suited to long-range relationship modeling. Dosovitskiy *et al.* [18] proposed the vision transformer (ViT) architecture that transforms 2-D image data into a 1-D sequence of tokens through splitting, linear projection, and flattening operations. The transformer encoder then replaces the standard CNN backbone for extracting deep features. Liu *et al.* [19] designed the Swin transformer, which is based on a shifted window scheme. Transformer-embedded structures have achieved state-of-the-art performances in multiple computer vision tasks, and their outstanding performance has also attracted the attention of the remote sensing community [12], [13].

---

[1] https://www.synapse.org/#!Synapse:syn3193805/wiki/217789
[2] https://www.cityscapes-dataset.com/
[3] http://host.robots.ox.ac.uk/pascal/VOC/voc2012/
[4] http://deepglobe.org

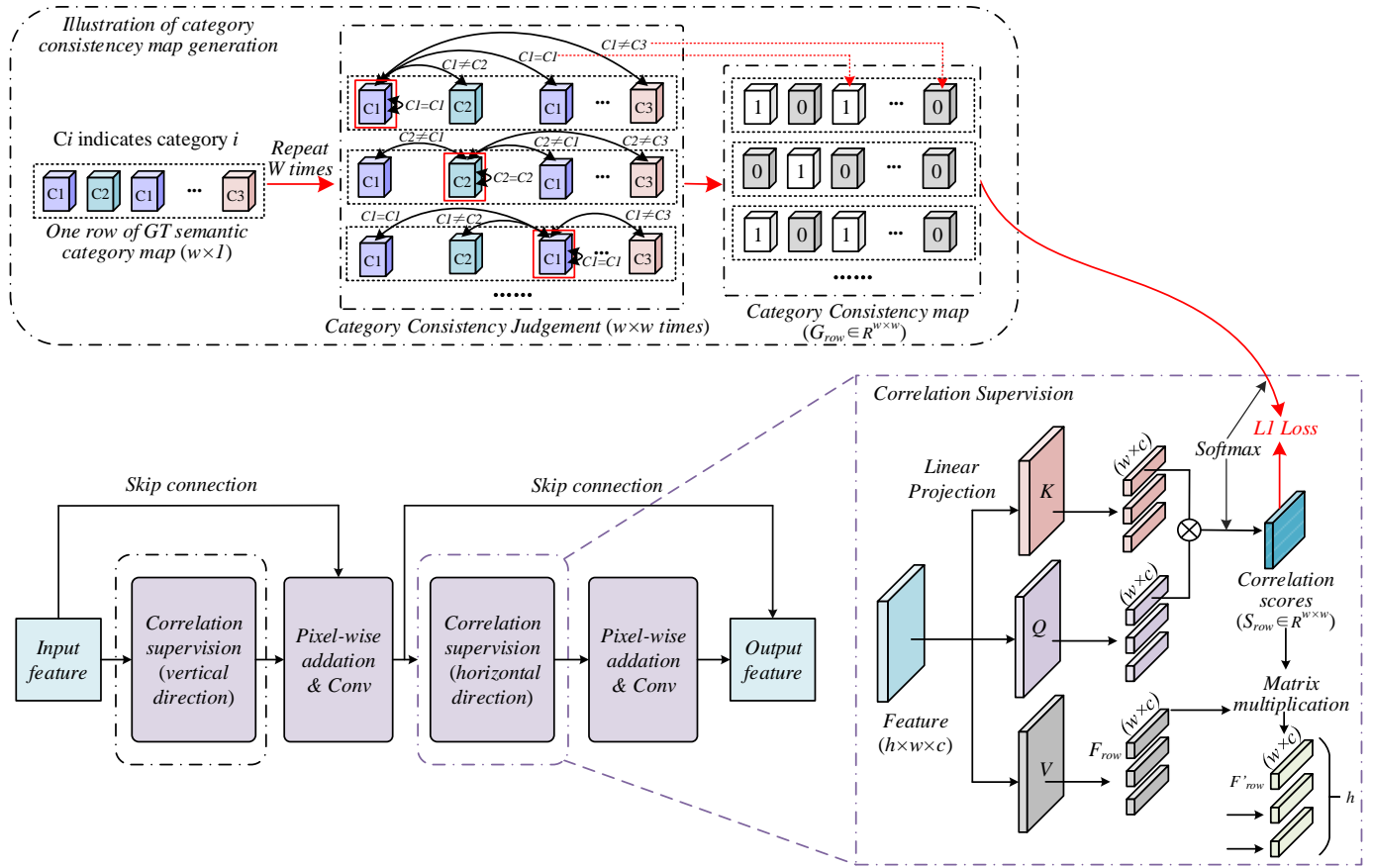

Fig. 2. The proposed long-range feature correlation supervision module (FCSM) and its supervised self-attention mechanism with category consistency constraints. The ground-truth (GT) category consistency map is computed based on the categories of two arbitrary pixels. 1 denotes they are the same and 0 different.

In the medical field, both the hybrid CNN and transformer architectures and pure transformer architectures have been considered [20], [21], [22]. Meanwhile, in the remote sensing field, several studies have found that pure transformer-based architectures obtain unsatisfactory performances because the transformer operation overly focuses on the global relationship modeling, but lacks fine positioning abilities [12], [13]. To cope with the complicated scenes in remote sensing images, the hybrid CNN and transformer architecture is more suitable [12], [13]. For instance, the Swin Transformer embedding U-Net (ST-UNet) [13] employs a dual encoder structure of the Swin transformer and a standard CNN in parallel to extract features, and the network proposed in [12] uses the Swin transformer as the encoder and the CNN as the decoder. In the encoder of the network [23], local features and global features are extracted by CNN and Transformer respectively, which are then fused in an interactive manner.

The long-range correlation modeling of the previous CNN-based attention mechanism and transformer structures are both effective. However, the unsupervised long-range correlation learning procedures in these models, i.e., there is only a series of matrix operations between local feature maps at different positions, without the supervision of any ground truth, can limit the ability of long-term dependency representation. In fact, each pixel/region has a clear category in the ground-truth labels of the semantic segmentation or LULC datasets, and an explicit learning goal can guide a more correct optimization direction than a self-optimization scheme. However, this information is not employed in long-range correlation modeling and has been ignored in previous studies.

In this paper, we propose a supervised long-range correlation method and introduce a corresponding efficient feature correlation supervision module (FCSM) to enhance the long-range information interaction capability of the semantic segmentation network. The FCSM decomposes the 2-D correlation into two 1-D correlations in the vertical and horizontal directions, respectively, and the computational complexity is significantly reduced. The semantic ground-truth maps are used as supervision information to recalibrate the intermediate features in the form of pixel-wise category consistency constraints. Thus, the proposed method helps the network produce more consistent segmentation results for pixels/regions in different positions but with the same category in remote sensing imagery.

*B. Multi-Scale Feature Extraction*

Multi-scale features are critical for the pixel-wise category prediction of remote sensing images that not only vary in ground resolution but also contain size-varied objects of different categories. In previous works, Li *et al.* [14] integrated self-smoothing Atrous convolutions with different radii to extract diverse features at multiple scales; Ji *et al.* [15] introduced two groups of parallel Atrous convolution layers with different dilation rates to extract multi-scale features; Deng *et al.* [16] embedded the Atrous spatial pyramid pooling module [24] into the backbone network to extract multi-scale features simultaneously; and Zhao *et al.* [25] designed a pyramid attention pooling module for adaptive feature refinement. Guan *et al.* [26] developed a multistage feature fusion lightweight model to fuse the multi-scale semantic feature maps from different stages of the encoder.

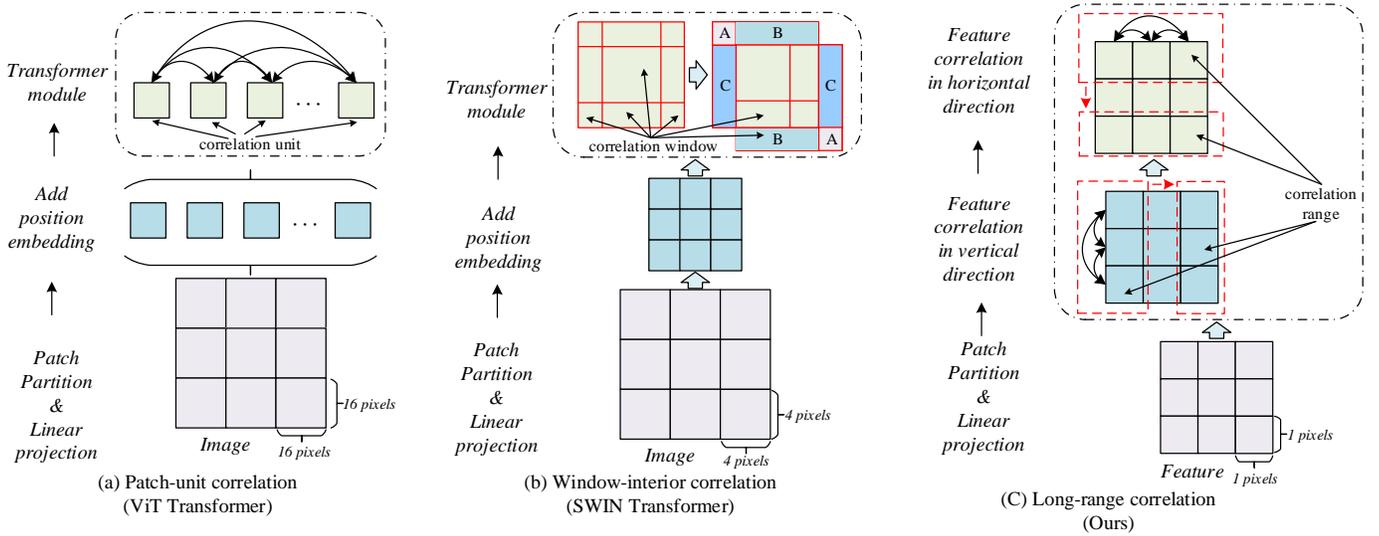

Fig. 3. The differences between the three correlation modeling strategies of the ViT, Swin transformer, and the proposed FCSM. The ViT conducts patch-unit based correlation on the whole 2-D input, the Swin transformer conducts window-interior correlation, and the proposed FCSM uses sequential vertical and horizontal correlations. The ViT/Swin transformer and the FCSM accept the original image and feature map as input, respectively, and 16, 4, and 1 is the pixel stride for the input used in the ViT, Swin transformer, and FCSM, respectively.

However, the above methods just extract multi-scale deep features from the input images of the original resolution. A CNN can capture richer detail features when it is fed with remote sensing images of more than one scale, providing a more comprehensive representation.

A few studies have applied this multi-scale input strategy. For example, Ji *et al.* [27] extracted deep features separately from two-scale inputs with a weight-shared U-Net; Ding *et al.* [28] proposed a CNN-based two-stage multi-scale training strategy, with the first stage extracting global features from downscaled images and the second stage extracting the local features from cropped original-resolution image patches; and Chen *et al.* [29] proposed GLNet, which is composed of a global branch and a local branch, taking the downsampled entire image and cropped original-resolution local patches as inputs, respectively, and fusing the corresponding features bidirectionally.

Although the works of [27]-[29] have utilized multi-scale image inputs, some drawbacks need to be addressed. For example, there are only two scales utilized for these three methods. Furthermore, the training procedure is not end-to-end in [28], which is inconvenient to implement. Most importantly, these methods extract global and local information from two-scale images, without considering the levels of the objects with various size differences. In fact, a more flexible approach to capturing comprehensive features of different-size objects from multi-scale images is preferable for segmenting remote sensing images.

In this work, we designed an auxiliary adaptive receptive field feature extraction module (ARFE) to enhance the size-adaptive capacity for the multi-size objects in multi-scale images. ARFE uses a switch mechanism to select suitable features generated from Atrous convolutions with different dilation rates, and it is applied to multi-scale downsampled images in a weight-unshared manner.

## III. METHODS

### A. Supervised Long-Range Correlation

In this paper, we propose a novel supervised long-range feature correlation method for LULC classification. The novelty lies in the explicit utilization of the category consistency constraints generated from the semantic ground-truth map to reduce the feature differences of the long-range regions with the same semantic category. The corresponding FCSM is designed to model the correlation of different regions in the feature map. The correlation scores (i.e., the weights for recalibration) are supervised by the category consistency constraints that enforce the recalibrated feature maps to have a better discriminative ability for classifying pixels of ambiguous categories.

The FCSM recalibrates the CNN features to achieve long-range dependency via a supervised self-attention mechanism and skip connections. The detailed workflow of the FCSM is shown in Fig. 2. The FCSM decomposes the 2-D correlation computation into two orthogonal 1-D correlation computations, which reduces the computational cost and GPU burden. The skip connection is employed to accelerate the model convergence. The pixel-wise addition and convolution

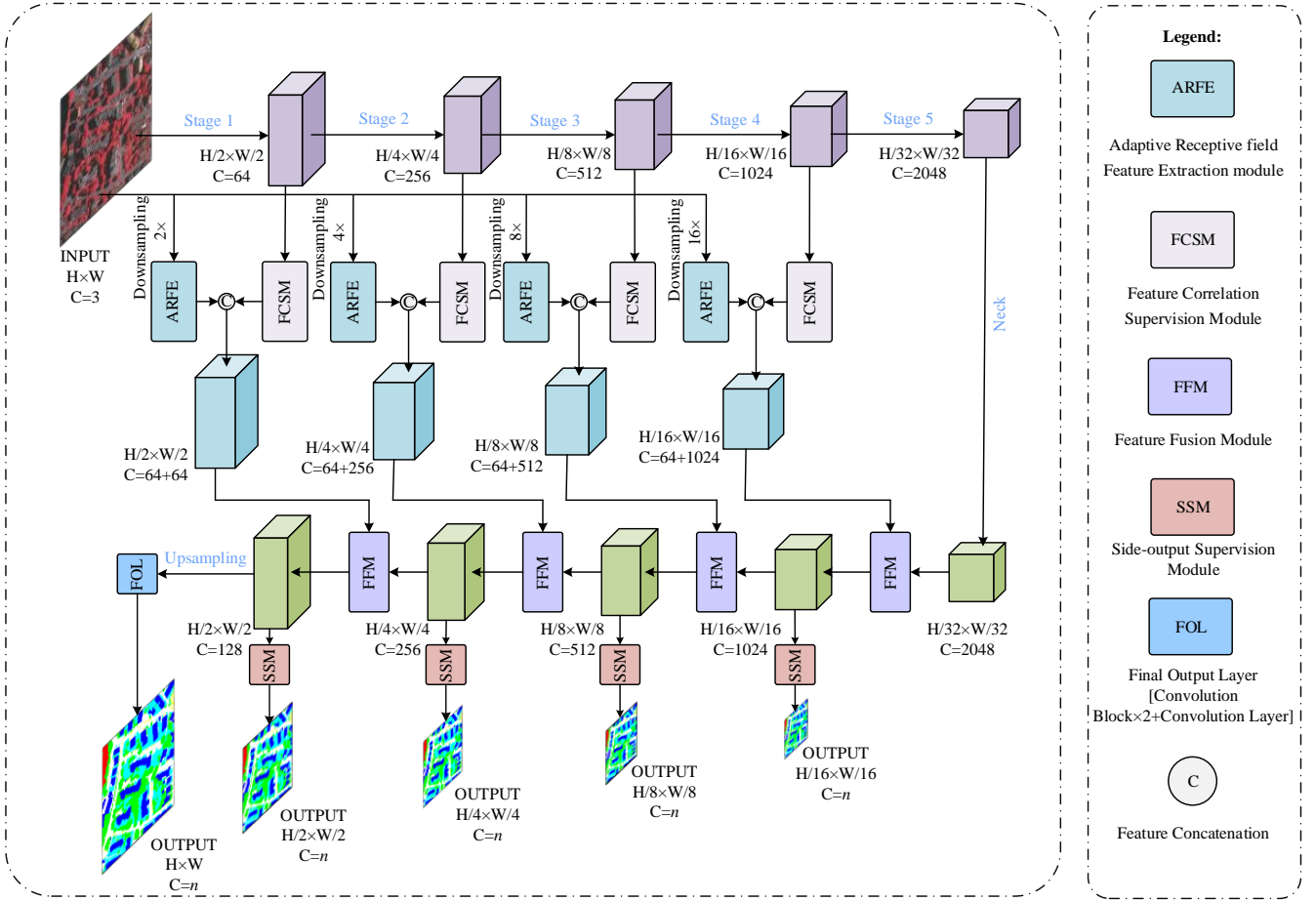

Fig. 4. The structure of SLCNet. Each adaptive receptive field feature extraction module is fed with a different downsampling-scale input image to produce a multi-scale feature description. H and W are the height and width of the input image, and *n* is the category number. Stages 1–5 are the ResNet-50 backbone, and the neck consists of two convolutional blocks.

operations are used to fuse the original input features and the optimized correlation features.

The core of the FCSM is the supervised self-attention mechanism, as shown in the enlarged part of Fig. 2. Given an input feature $F_{in} \in R^{H \times W \times C}$, three trainable linear projections map the input to three different feature vectors, named $K \in R^{H \times W \times C}$ (key), $Q \in R^{H \times W \times C}$ (query), and $V \in R^{H \times W \times C}$ (values), respectively. The output features are then computed as a weighted sum of $V$, where the weight assigned to each value in $V$ is computed by a compatibility function of $Q$ with the corresponding $K$ [30]. In our case study, the self-attention operation was conducted by row and column, respectively, instead of performing correlations between each pixel on the entire feature map.

The score calculated from the features at position $P_K$ in the Key ($K$) and position $P_Q$ in the query ($Q$) reflects their correlation [30]. A higher correlation score can be expected when the categories of the two positions are consistent, and vice versa. $P_Q$ and $P_V$ are located on the same row or column, and the correlation supervision is naturally extended to the entire 2-D feature map after the vertical and horizontal calculations are performed sequentially. We then introduce the supervision information based on the category consistency constraints, which is a binary map obtained from the ground truth (where 0 indicates different categories and 1 indicates the same category), which explicitly pushes the feature vectors with the same category close in the $C$-dimension feature space, no matter how great the distances between the features are.

Taking the correlation process in the horizontal direction as an example, a row of the value ($V$) feature vector are denoted as $F_{row} \in R^{w \times C}$ where $w$ is the width of the vector and $C$ is the channel number. Through self-attention operations using the $Q$ and $V$ feature vectors, we obtain correlation scores $S_{row} \in R^{w \times w}$, which represent the correlation degree of pixels at different positions. We supervise the correlation scores with the ground truth category consistency constraint map where pixels sharing the same category have the high consistency scores and pixels from different categories have low scores. We employ the L1 loss to optimize the correlation scores. The row and column features are recalibrated respectively, as shown in Eq. (1) and (2), to improve their discriminative ability for classifying pixels of ambiguous categories.

$$F'_{row} = S_{row} \times F_{row} \qquad (1)$$

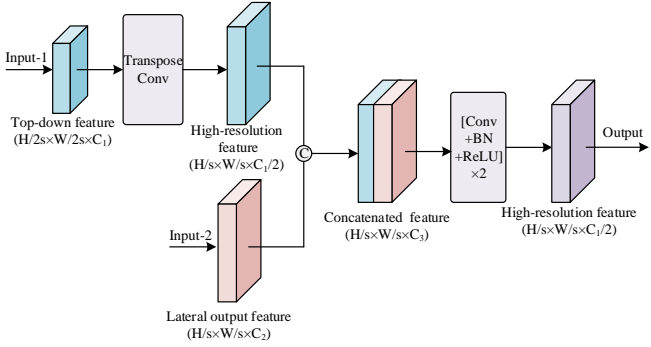

Fig. 5. The feature fusion module (FFM) in the proposed SLCNet.

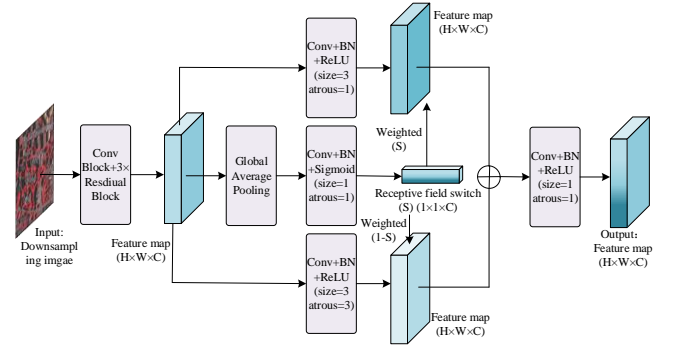

Fig. 6. The proposed adaptive receptive field feature extraction (ARFE) module.

$$F'_{col}=S_{col} \times F_{col} \qquad (2)$$

where "×" is matrix multiplication, $F'_{row} \in R^{w \times C}$ and $F'_{col} \in R^{h \times C}$. The corresponding losses are calculated as follows:

$$Loss_{row}=\frac{1}{w \times w}\sum_{i=1}^{i=w}\sum_{j=1}^{j=w}L_1(S_{ij},G_{ij}) \qquad (3)$$

$$Loss_{col}=\frac{1}{h \times h}\sum_{i=1}^{i=h}\sum_{j=1}^{j=h}L_1(S_{ij},G_{ij}) \qquad (4)$$

where $h$ and $w$ is the height and width of the value ($V$) feature vector, $i$ and $j$ are indexes of arbitrary pixel pairs. As shown in Fig. 2, both the correlation score $S$ and its ground truth $G$ in formulas (3) and (4) have been normalized using the softmax function. The total loss of FCSM is computed by averaging the losses of rows and columns:

$$Loss_{FCSM}=\frac{1}{h}\sum_{i=1}^{i=h}Loss_{row}+\frac{1}{w}\sum_{i=1}^{i=w}Loss_{col} \qquad (5)$$

Note that the FCSM can be regarded as a new variant of an attention module when supervision information is not used. We found that the unsupervised FCSM is still more effective than the ViT and Swin transformer when segmenting remote sensing images. The ViT conducts self-attention operations in all the tokens of the input once, which results in a huge GPU memory cost. Therefore, it only uses low-resolution feature maps (with a 16× downsampling stride to the input images) during the whole encoding stage. The Swin transformer restricts the 2-D self-attention computation into local small windows to reduce the computational complexity, but the explicit correlation range is shorted. Differing from the ViT and Swin transformer, the proposed FCSM, which is based on the sequentially connected vertical and horizontal 1-D correlation modeling, can process high-resolution feature maps, with much less GPU cost; meanwhile, the explicit correlation range is still long. This is of key importance for a large-capacity remote sensing image. The differences between the three correlation modeling strategies are clearly shown in Fig. 3. The ViT and Swin transformer take 16× and 4× downsampled images as input, while the proposed FCSM processes a feature map with the original scale.

### B. The Proposed Network

To evaluate the effectiveness of the proposed supervised long-range modeling method, we integrated it into a popular encoder-decoder style network, which we call the supervised long-range correlation network (SLCNet). In addition to FCSM, SLCNet has another module called the adaptive receptive field feature extraction module (ARFE) and a side-output supervision module (SSM) for enhancing the multi-scale feature representation ability.

The structure of SLCNet is illustrated in Fig. 4. In the encoder, ResNet-50 [31] is utilized to process the original-scale image. The feature maps are progressively deepened and downsampled, and passed to FCSM for global long-range information interaction. Simultaneously, ARFE takes the 2×, 4×, 8×, and 16× downsampled images as inputs, extracting detailed features for objects of various sizes. The size-adaptive features obtained from ARFE and the recalibrated global features from FCSM are then concatenated at their corresponding scales. The long-range feature correlation is not conducted on the 32× downsampled feature map as it already has a large enough receptive field to capture the global context information. In the decoder stage, a simple feature fusion module (FFM, see Fig. 5) upsamples the top-down features by transposed convolution, concatenates the upsampled features with the lateral features, and then applies a convolutional layer to refine the information. The multi-scale side outputs are compared with the downsampled ground-truth maps to provide auxiliary supervision information. The final pixel-wise category decision is made by picking the class with the maximum score along the channel dimension from the full-resolution output.

ARFE, side-output supervision and the hybrid loss function adopted in the training procedure are described in detail in the following sections.

### C. The Adaptive Receptive Field Feature Extraction Module

The designed ARFE produces adaptive and detailed feature descriptions of the multi-size objects in the multi-scale images via a simple and lightweight structure. Each ARFE captures multi-size features at the current scale, and multiple ARFEs work on the different-scale images in a weight-unshared manner. In particular, a switch mechanism at each ARFE is employed to flexibly control the generated feature description, as shown in Fig. 6.

ARFE takes a downsampled image as input and outputs features of the same resolution. Firstly, a convolutional block containing a convolutional layer, a batch normalization layer, a rectified linear unit (ReLU) activation layer, and three standard residual blocks [31] encodes the input image to $C$-dimension general (i.e., size-insensitive) features. The three subsequent parallel branches are designed to collaboratively enhance the features with better size sensitivity, forming the kernel of ARFE. The top branch and the bottom branch are convolutional blocks with different dilation rates, where the top branch has a small dilation rate and the bottom branch has a large dilation rate, for adequately describing small objects and large objects, respectively. The middle branch is a switch generator consisting of a global average pooling layer and a convolutional block with sigmoid activation function, which generates a receptive field switch weight $S \in R^{1 \times 1 \times C}$. $S$ weights the small receptive field features, and thus $1-S$ weights the large receptive field features. The weighted features from the top and bottom branches are pixel-by-pixel added and then convolved by the final convolutional block. Please note that we apply channel-wise switch weights instead of pixel-wise ones because the different channels of feature maps already contain discriminative information about the objects of different categories [32], and pixel-wise weights are costly and difficult to learn.

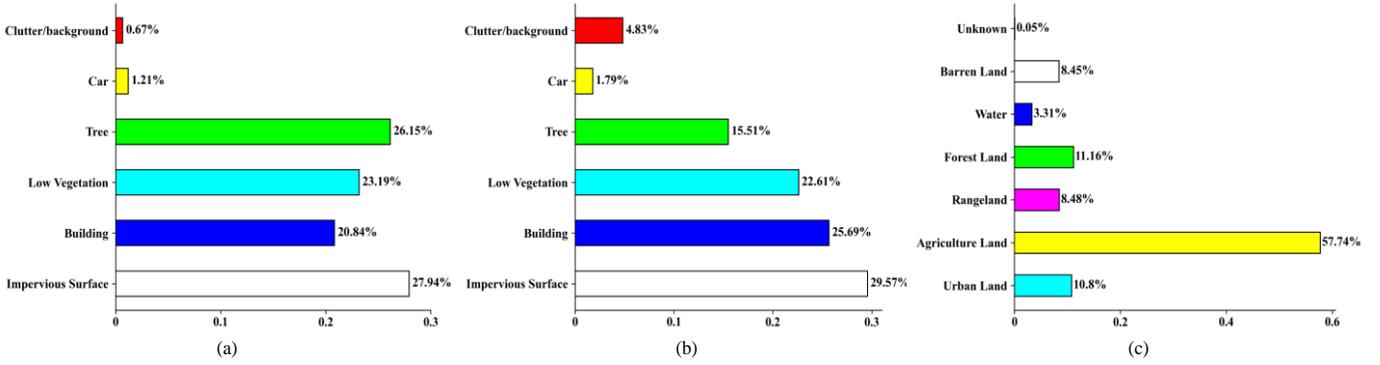

Fig. 7. The proportion of each semantic category in the Vaihingen dataset (a), the Potsdam dataset (b), and the DeepGlobe dataset (c).

### D. Side-Output Supervision and Hybrid Loss

*1) Side-Output Supervision Module:* To make the SLCNet better learn multi-scale features, the SSM is adopted at multiple scales. Each side-output layer consists of a convolutional block and a convolutional layer, which maps the high-dimension features to the category prediction maps. At each scale, the output is supervised by the downsampled ground-truth semantic category map obtained via the nearest neighbor interpolation method. *2) Hybrid Loss:* The loss function employed during the training stage estimates the difference between the prediction maps from the CNN and the manually labeled ground-truth maps, providing directions to update the model parameters by backpropagation. Multiple-type loss functions can be applied collaboratively to achieve better supervision.

The widely used logistic regression with cross-entropy (CE) loss is a relatively poor indicator of the quality of segmentation. It measures the per-pixel prediction error with evaluation bias because it naturally focuses more on the categories that dominate in pixel quantity. Lovász softmax (LS) loss [33] provides a global perspective to measure the error of the entire prediction map compared to the ground-truth map, regardless of the imbalance of the different categories. The LS loss is based on the Jaccard index (which is sometimes called the intersection over union score or IoU score). As a result, the LS loss is beneficial for recognizing minor categories with few pixels in the images. We observed that CE loss and LS loss are complementary, and their combination can form a collaborative loss function addressing both the details and the whole. The aim is to obtain a high IoU score for the final prediction map and fine details for each downscaled prediction map, so the LS loss function is applied in the final output layer and the CE loss function is applied in the side-output layers.

As mentioned above, we employed $L_1$ loss to reduce the differences of the features belonging to the same category in the FCSM. Therefore, the total loss function of the proposed method is as follows:

$$Loss_{all} = \alpha \times Loss_{FCSM} + \beta \times Loss_{Side\ outputs} + \gamma \times Loss_{Final\ output} \quad (6)$$

where $\alpha$, $\beta$, and $\gamma$ respectively weigh the $L_1$ loss for the FCSM, the CE loss for the side-output layers, and the LS loss for the final output layer. After trial and error, we set $\alpha = 10$, $\beta = 0.05$, and $\gamma = 1$ in this study.

The losses of side-output layers and the final output layer in SLCNet are calculated as follows:

$$Loss_{Side\ outputs} = \sum_{s=1}^{s=4} CE(M_{pt,s}, M_{gt,s}) \quad (7)$$

$$Loss_{Final\ output} = LS(M_{pt}, M_{gt}) \quad (8)$$

where $M_{pt,s}$ and $M_{gt,s}$ represent the predicted map and the ground truth on the scale of $s$.

## IV. EXPERIMENTS AND RESULTS

In this section, we describe how the proposed SLCNet and the other state-of-the-art semantic segmentation methods from the computer vision, medicine, and remote sensing communities were compared. Three publicly available datasets were employed to conduct the experiments, i.e., the ISPRS Vaihingen and Potsdam aerial datasets and the DeepGlobe land cover classification satellite dataset. Details of the three datasets are provided in Section III-A; the evaluation metrics and the experimental settings are provided in Section III-B; the comparison results are presented in Section III-C; and the ablation experiments conducted for each part of the proposed method and discussions about the pretrained weights and model

TABLE I
COMPARISON OF DIFFERENT METHODS ON THE VAIHINGEN DATASET

| Method | Backbone | Per-Class IoU Score (%)/Per-Class F1 Score (%) | | | | | mIoU (%) | ave.F1 (%) |
| --- | --- | --- | --- | --- | --- | --- | --- | --- |
| | | Imp.surf | Building | Low veg | Tree | Car | | |
| U-Net [41] | ResNet-50 | 79.2/88.4 | 87.8/93.5 | 60.6/75.4 | 74.2/85.2 | 69.1/81.7 | 74.2 | 84.9 |
| PSPNet [38] | ResNet-50 | 79.4/88.5 | 87.2/93.1 | 62.1/76.6 | 74.3/85.3 | 65.9/79.5 | 73.8 | 84.6 |
| RefineNet [39] | ResNet-50 | 78.7/88.1 | 87.4/93.2 | 60.4/75.3 | 73.4/84.7 | 70.1/82.4 | 74.0 | 84.7 |
| DeepLabv3+ [40] | ResNet-50 | 77.9/87.6 | 87.1/93.1 | 59.9/74.9 | 73.6/84.8 | 60.0/75.0 | 71.7 | 83.1 |
| DANet [17] | ResNet-50 | 77.4/87.3 | 86.3/92.6 | 60.9/75.7 | 73.0/84.4 | 54.8/70.8 | 70.5 | 82.2 |
| Swin-UNet [20] | Swin-Tiny | 79.1/88.3 | 86.4/92.7 | 62.3/76.8 | 74.1/85.1 | 63.8/77.9 | 73.1 | 84.2 |
| Trans-UNet [21] | R50-ViT-Base | 80.4/89.2 | **88.2/93.7** | 62.4/76.9 | 74.6/85.4 | 70.2/82.5 | 75.2 | 85.5 |
| ST-UNet [13] | R50-ST | 78.7/88.1 | 86.2/92.6 | 61.3/76.0 | 73.9/85.0 | 63.9/77.9 | 72.8 | 83.9 |
| SLCNet (Proposed) | ResNet-50 | **81.1/89.6** | 88.1/93.6 | **64.5/78.4** | **75.0/85.7** | **74.6/85.5** | **76.7** | **86.6** |

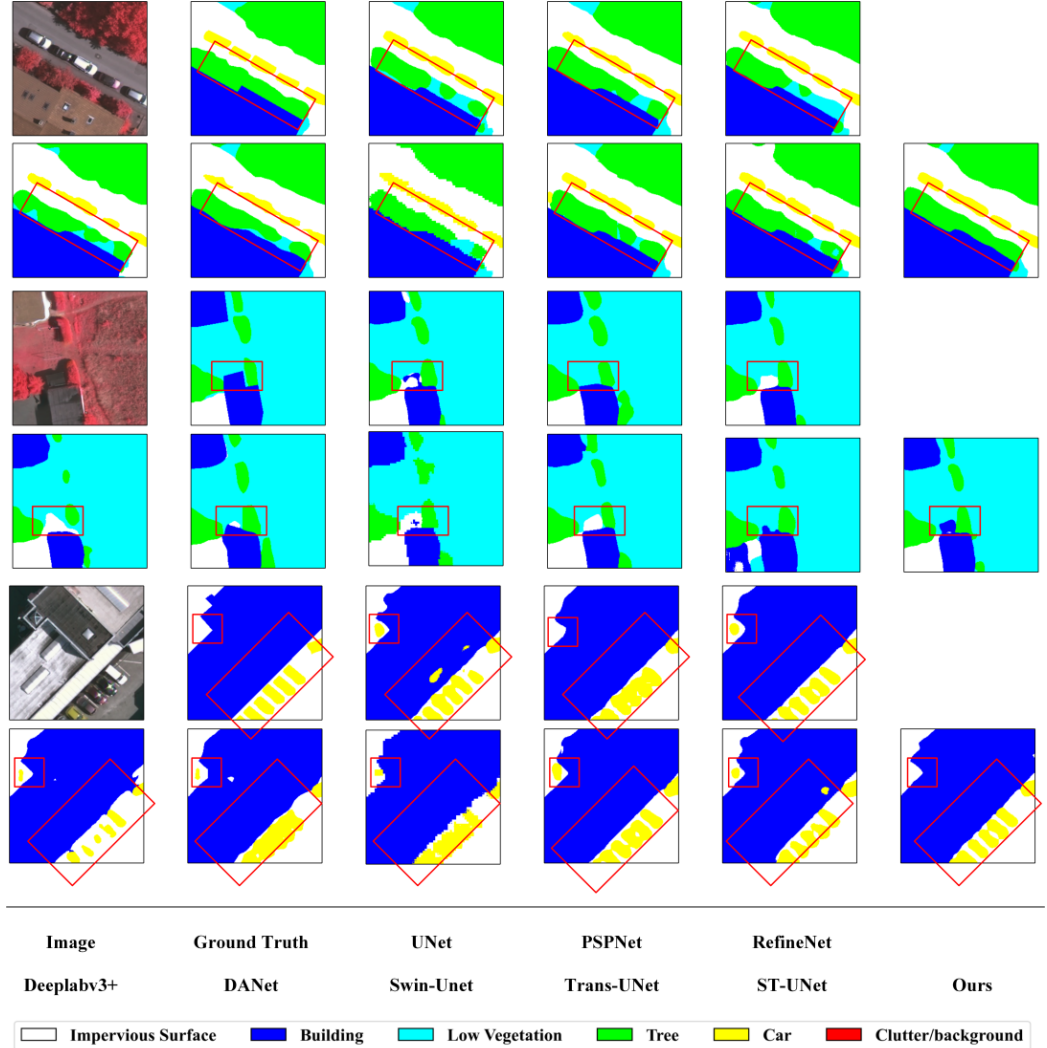

Fig. 8. Examples of the segmentation results obtained on the Vaihingen test set by the different methods. The layout sequence of original image, ground truth, and results is listed according to the words below the black line. The red rectangles indicate the error-prone regions in the images.

efficiency are provided in Sections III-D–III-E.

A. Datasets

1) Vaihingen Dataset: The ISPRS Vaihingen dataset [34] contains 33 true orthophoto images (red, green, and near-infrared bands) generated by aerial photogrammetry technology. The ground sampling distance (GSD) of the images is around 9 cm, and each image size varies from 1996 × 1995 pixels to 3816 × 2550 pixels. Six land-cover categories are labeled for semantic segmentation, i.e., impervious surface (Imp. surf), building, low vegetation (Low veg), tree, car, and clutter/background. Following the works of [13], [35], we used 11 images for the training and five images for the testing. The training image IDs are 1, 3, 5, 7, 13, 17, 21, 23, 26, 32, and 37. The test image IDs are 11, 15, 28, 30, and 34. The proportion of each semantic category in the Vaihingen dataset is shown in

TABLE II
COMPARISON OF THE DIFFERENT METHODS ON THE POTSDAM DATASET

| Method | Backbone | Per-Class IoU Score (%)/Per-Class F1 Score (%) | | | | | mIoU (%) | ave.F1 (%) |
|---|---|---|---|---|---|---|---|---|
| | | Imp.surf | Building | Low veg | Tree | Car | | |
| U-Net [41] | ResNet-50 | 86.4/92.7 | 94.4/97.1 | 74.6/85.4 | 73.7/84.9 | 85.6/92.2 | 82.9 | 90.5 |
| Trans-UNet [21] | R50-ViT-Base | 87.1/93.1 | 93.7/96.8 | 76.3/86.5 | 76.1/86.4 | 84.6/91.7 | 83.6 | 90.9 |
| SLCNet (proposed) | ResNet-50 | **88.5/93.9** | **95.1/97.5** | **77.7/87.4** | **77.0/87.0** | **87.5/93.4** | **85.2** | **91.8** |

TABLE III
COMPARISON OF THE DIFFERENT METHODS ON THE DEEPGLOBE DATASET

| Method | Backbone | Per-Class IoU Score (%)/Per-Class F1 Score (%) | | | | | | mIoU (%) | ave.F1 (%) |
|---|---|---|---|---|---|---|---|---|---|
| | | Urban land | Agriculture land | Rangeland | Forest land | Water | Barren land | | |
| U-Net [41] | ResNet-50 | 76.0/86.4 | 85.0/91.9 | 34.0/50.8 | 77.5/87.3 | 80.4/89.1 | 55.5/71.4 | 68.1 | 79.5 |
| Trans-UNet [21] | R50-ViT-Base | 77.9/87.6 | 86.8/92.9 | 36.5/53.5 | 76.6/86.8 | 83.6/91.1 | 59.8/74.8 | 70.2 | 81.1 |
| SLCNet (proposed) | ResNet-50 | **78.1/87.7** | **87.8/93.5** | **40.1/57.2** | **79.9/88.8** | **83.9/91.2** | **62.8/77.1** | **72.1** | **82.6** |

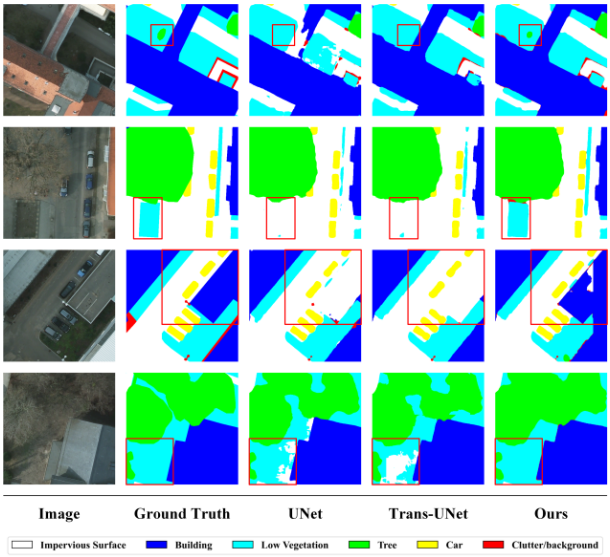

Fig. 9. Examples of the segmentation results of the different methods obtained on the Potsdam test set. The red rectangles indicate the error-prone regions in the images.

Fig. 7(a).

*2) Potsdam Dataset:* The ISPRS Potsdam dataset [34] contains 38 true orthophoto images. Images with red, green, blue, and near-infrared spectral bands are provided. Referring to [13], we used the images with red, green, and blue bands for training and evaluation in the experiments. The GSD of the images in the Potsdam dataset is around 5 cm, and each image is of 6000 × 6000 pixels in size. Each pixel is labeled to one of the six land-cover categories, as with the Vaihingen dataset. Following the previous works [10], [13], we used 24 images for training and the remaining 14 images for testing. The test image IDs are 2_13, 2_14, 3_13, 3_14, 4_13, 4_14, 4_15, 5_13, 5_14, 5_15, 6_13, 6_14, 6_15, and 7_13. The proportion of each semantic category in the Potsdam dataset is shown in Fig. 7(b).

*3) DeepGlobe Dataset:* The DeepGlobe land cover classification dataset [36] provides 803 sub-meter resolution satellite images and annotated maps. Each image in this dataset has a size of 2448 × 2448 pixels, and the GSD is about 50 cm. Following the work of [37], we randomly selected 602 images for the training and the other 201 were used for the performance evaluation. Differing from the Vaihingen dataset (which covers a small village) and the Potsdam dataset (which covers a historic city), the DeepGlobe dataset focuses more on rural areas. Seven land-cover types are labeled in the DeepGlobe dataset, i.e., urban land, agriculture land, rangeland, forest land, water, barren, and unknown. The lower GSD and the more complex land-cover types make the dataset more challenging than the Vaihingen and Potsdam datasets. The proportion of each semantic category in the DeepGlobe dataset is shown in Fig. 7(c).

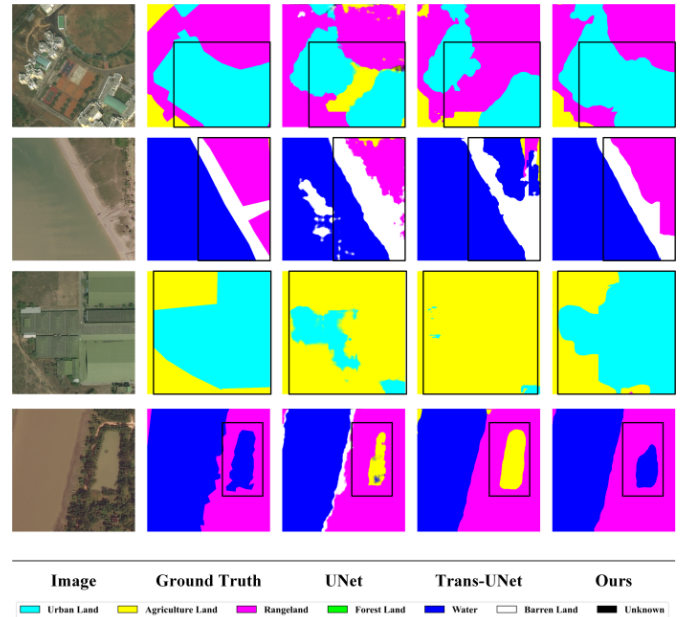

Fig. 10. Examples of the segmentation results obtained on the DeepGlobe test set. The black rectangles indicate the error-prone regions in the images.

*B. Evaluation Metrics and Experimental Settings*

*1) Evaluation Metrics:* The IoU score and F1 score for each category and the mean intersection over union (mIoU) score and average F1 score (ave.F1) for all the categories are employed to evaluate the performance of the proposed SLCNet and all of the comparison methods. For each semantic category, the IoU and F1 scores are defined as follows:

$$IoU = \frac{\|TP\|}{\|TP\| + \|FP\| + \|FN\|} \quad (9)$$

TABLE IV
QUANTITATIVE RESULTS OF INTRODUCING DIFFERENT MODULES IN THE PROPOSED METHOD ON THE VAIHINGEN DATASET

| Baseline | SSM | Hybrid Loss | ARFE | FCSM | UFCM | mIoU (%) | ave.F1 (%) | Params (M) |
|---|---|---|---|---|---|---|---|---|
| √ | | | | | | 74.2 | 84.9 | 147.8 |
| | | *Output supervision* | | | | | | |
| √ | √ | | | | | 74.8 | 85.3 | 148.9 |
| √ | √ | √ | | | | 75.0 | 85.5 | 148.9 |
| | | | *Functional module* | | | | | |
| √ | √ | √ | √ | | | 75.8 | 86.0 | 151.2 |
| √ | √ | √ | | √ | | 76.2 | 86.2 | 184.8 |
| √ | √ | √ | √ | √ | | **76.7** | **86.6** | 187.1 |
| | | | *Correlation supervision* | | | | | |
| √ | √ | √ | √ | √ | | **76.7** | **86.6** | 187.1 |
| √ | √ | √ | √ | | √ | 76.2 | 86.2 | 187.1 |

TABLE V
QUANTITATIVE RESULTS OF INTRODUCING DIFFERENT CORRELATION MODULES IN THE PROPOSED METHOD ON THE VAIHINGEN DATASET
(NEW BASELINE = BASELINE + SSM + ARFE + HYBRID LOSS)

| Method | mIoU (%) | Ave.F1 (%) | Params (M) |
|---|---|---|---|
| New baseline | 75.8 | 86.0 | 151.2 |
| New baseline +UFCM | 76.2 | 86.2 | 187.1 |
| New baseline + VTM | 75.6 | 85.9 | 200.7 |
| New baseline + STM | 75.9 | 86.0 | 198.3 |
| New baseline + FCSM | **76.7** | **86.6** | 187.1 |

$$F1 = \frac{2 \times Precision \times Recall}{Precision + Recall} \quad (10)$$

where *TP* means the true positive pixels in the prediction maps, *FP* is the false positive pixels, and *FN* is the false negative pixels. The $\|\cdot\|$ symbol refers to counting the number of all the *TP* (or *FP*, *FN*) pixels. The precision and recall in the formula (3) are computed as follows:

$$Precision = \frac{\|TP\|}{\|TP\| + \|FP\|} \quad (11)$$

$$Recall = \frac{\|TP\|}{\|TP\| + \|FN\|} \quad (12)$$

The mean intersection over union score (mIoU) and the average F1 score (ave.F1) are obtained by computing the arithmetic average values of the IoU scores and F1 scores of all the categories.

*2) Experimental Settings:* For a fair comparison, all the experiments were executed on a Windows PC equipped with an NVIDIA TITAN RTX 24 GB GPU and an Intel Core i9-9900K CPU, and all the comparison methods were implemented in the PyTorch deep learning framework. Due to the memory limitation of the GPU, we cropped the original images/ground-truth maps into small-size image tiles in an offline manner when preparing the training dataset. In particular, we cropped the images to $256 \times 256$-pixel tiles for the small Vaihingen dataset to keep the same experimental configuration as the very recent work [13]. Cutting images into smaller tiles will bring more combinations of mini-batch during the training procedure. For the Potsdam dataset and the DeepGlobe dataset, the size of the cropped tiles was $512 \times 512$ pixels because significantly more training images were available. For each dataset, 10% of the cropped tiles were randomly selected from the training set as the sequestered validation set.

During the training stage, the initial learning rate was set to 1e-4 and divided by 10 at every 33/13/13 epochs. The maximum number of training epochs was set to 100, 40, and 40 for the Vaihingen dataset, Potsdam dataset, and DeepGlobe dataset, respectively. The Adam optimizer was used to update the model parameters. Data augmentation strategies, including histogram equalization, Gaussian filtering, and random horizontal and vertical flipping, were applied for training all the methods. During the testing stage, the original image was cropped into small-size tiles at a 1/4 overlap rate in an online manner and fed into the networks. Their prediction results were then merged to produce a complete prediction map with the same size as the original large image.

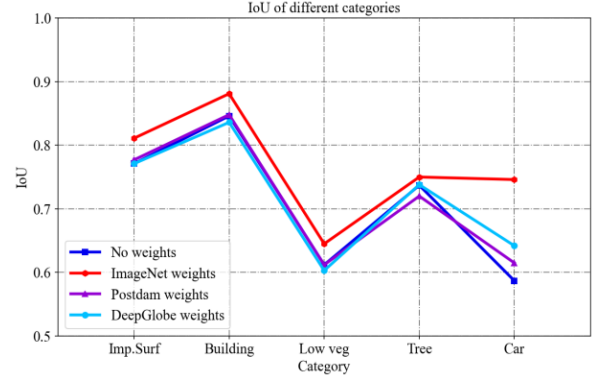

Fig. 11. IoU score of each category on the Vaihingen dataset when initializing the proposed SLCNet method with different pre-trained weights.

Following the works of [12], [13], [29], we ignored the category of *clutter/background* in the Vaihingen dataset and Potsdam dataset, and also the category of *unknown* in the DeepGlobe dataset when performing the performance evaluation.

*C. Comparison with the State-of-the-Art Methods*

*1) Vaihingen Dataset:* We compared the proposed SLCNet method with the recent semantic segmentation methods from the computer vision, medicine, and remote sensing communities on the Vaihingen dataset with the same training-validation-testing data division and training procedure. PSPNet [38], RefineNet [39], DeepLabv3+ [40], and DANet [17] are from the computer vision community for semantic segmentation of close-range natural images; U-Net [41], Swin-UNet [20], and Trans-UNet [21] were developed in the medical field for segmenting medical images; ST-UNet [13] is a recently proposed remote sensing image segmentation method; Swin-UNet [20] and ST-UNet [13] introduce the Swin transformer encoder for long-range modeling; and Trans-UNet [21] employs the ViT encoder to enhance the context extraction capability. We found that the ST-UNet model we trained with ImageNet pretrained weights [42] obtained a better segmentation performance (mIoU score of 72.8%) than the authors reported (mIoU score of 70.2%) in their original work [13] (we modified the half-compressed ResNet-50 backbone to an uncompressed one for loading the ImageNet weights). To make the comparison fair, all the comparison methods and the proposed SLCNet method were initialized using the ImageNet pretrained weights. All the comparison methods use encoders of a similar scale, except for Swin-UNet [20], because its authors only released the Swin-Tiny-encoder based version and declared that an increase of the model scale would hardly improve the performance. Therefore, the parameters of Swin-UNet [20] are significantly less than those of the other methods (see Table VIII).

TABLE VII
QUANTITATIVE RESULTS OF THE PROPOSED METHOD ON THE VAIHINGEN DATASET WHEN USING DIFFERENT PRE-TRAINED WEIGHTS

| Retraining/test dataset | No pre-training weights (mIoU/ave.F1) | ImageNet weights (mIoU/ave.F1) | Potsdam dataset weights (mIoU/ave.F1) | DeepGlobe dataset weights (mIoU/ave.F1) |
|---|---|---|---|---|
| Vaihingen dataset | 71.1/82.6 | **76.7/86.6** | 71.5/83.0 | 71.8/83.3 |

TABLE VI
QUANTITATIVE RESULTS OF THE PROPOSED METHOD WITH AND WITHOUT THE IMAGENET PRE-TRAINED WEIGHTS ON THE THREE DATASETS

| Dataset | w/ pre-training weight (mIoU/ave.F1) | w/o pre-training weight (mIoU/ave.F1) |
|---|---|---|
| Vaihingen dataset | **76.7/86.6** | 71.1/82.6 |
| Potsdam dataset | **85.2/91.8** | 77.4/87.0 |
| DeepGlobe dataset | **72.1/82.6** | 65.5/77.3 |

The quantitative evaluation results of all the methods are listed in Table I. The proposed SLCNet obtains both the highest mIoU of 76.7% and the highest ave.F1 of 86.6%, outperforming the second-best Trans-UNet [21] by 1.5% and 1.1%, respectively. At the category level, the proposed SLCNet also performs excellently. SLCNet achieves the highest IoU and F1 scores in four out of five categories, whereas the second-best Trans-UNet method only leads in one category. The results demonstrate the effectiveness of the newly introduced supervised long-range correlation, ARFE, and hybrid loss in SLCNet.

Several examples of the segmentation results from the Vaihingen test set are shown in Fig. 8. The red rectangles indicate the error-prone regions in the images. In the first row of Fig. 8, all the comparison methods mistake parts of the trees for low vegetation, but SLCNet obtains consistent segmentation results in these tree regions. In the second row of Fig. 8, shadows make the recognition difficult, but the proposed method still segments the building regions better than the other methods. In the last row of Fig. 8, SLCNet performs well in distinguishing the cars and buildings/shadows sharing similar textures and colors. Another observation is that Swin-UNet always produces rough-edged segmentation results, which is because this pure transformer structure focuses too much on modeling the long-range correlation and lacks fine positioning capabilities.

TABLE VIII
PARAMETERS AND EFFICIENCIES OF THE DIFFERENT METHODS ON THE VAIHINGEN TEST SET

| Method | Params (M) | EFFICIENCY (seconds) |
|---|---|---|
| U-Net [41] | 147.8 | 11.8 |
| PSPNet [38] | 48.9 | 10.9 |
| RefineNet [39] | 94.9 | 10.4 |
| DeepLabv3+ [40] | 40.3 | 8.0 |
| DANet [17] | 49.5 | 8.9 |
| Swin-UNet [20] | 27.2 | 12.3 |
| Trans-UNet [21] | 107.5 | 18.3 |
| ST-UNet [13] | 299.3 | 27.1 |
| SLCNet (Proposed) | 187.1 | 16.0 |

*2) Potsdam Dataset:* For the Potsdam dataset and the DeepGlobe dataset, we selected U-Net [41], which outperformed the other standard CNN segmentation methods on the Vaihingen dataset, and Trans-UNet [21], which outperformed the other recent transformer-boosted segmentation methods. The quantitative results of U-Net [41], Trans-UNet [20], and SLCNet are listed in Table II. Due to the larger data size, all three methods obtain higher IoU and F1 scores than on the Vaihingen dataset. The proposed SLCNet achieves the best IoU and F1 scores in all five categories, and outperforms the second-best Trans-UNet by 1.6% and 0.9% in terms of mIoU and ave.F1.

Several segmentation maps obtained on the Potsdam test set by the three methods are shown in Fig. 9. SLCNet shows an advantage in distinguishing low vegetation and tree (the first row), low vegetation and impervious surface (the second and fourth rows), and building and impervious surface (the third row), which have similar colors and textures that heavily impact the other methods. Trans-UNet, although enhanced with long-range dependency modeling, shows little difference with the plain U-Net, while SLCNet obtains significantly better segmentation maps. This demonstrates that the supervised long-range correlation performs better than the self-optimized correlation in terms of helping the network segment the ambiguous regions correctly.

*3) DeepGlobe Dataset:* The satellite images in the DeepGlobe dataset have a lower ground resolution, and thus all three methods obtain lower IoU and F1 scores than on the Vaihingen dataset and Potsdam aerial dataset. From Table III, it can be seen that SLCNet obtains the best IoU and F1 scores in all six categories and outperforms the second-best Trans-UNet by 1.9% and 1.5% in terms of mIoU and ave.F1. U-Net obtains the worst performance in four of six categories and the comprehensive mIoU and ave.F1.

Visualization results of the different methods obtained on the

> REPLACE THIS LINE WITH YOUR PAPER IDENTIFICATION NUMBER (DOUBLE-CLICK HERE TO EDIT) <    12DeepGlobe test set are shown in Fig. 10. For the first row, U-Net and Trans-UNet incorrectly classify the sports ground in the urban land into agriculture land and rangeland respectively, as labeled by the black rectangles. In contrast, the proposed SLCNet achieves more consistent segmentation results in this urban land. Similar confusion also happens in the rangeland category in the second row, the urban land category in the third row, and the water category in the fourth row. The explicit long-range correlation supervision can help the network predict consistent classification results in regions of the same semantic category, resulting in SLCNet performing better than those methods without this strategy.

*D. Effectiveness of Individual Parts*

To evaluate the effectiveness of each newly introduced component in SLCNet, we conducted incremental ablation experiments. Taking U-Net with ResNet-50 backbone as the baseline, we progressively tested the effectiveness of different components: the output supervision, the functional modules, and the correlation supervision mechanism. As shown in Table IV, the introduction of side-output supervision module (SSM) improved the baseline's performance by 0.6%/0.4% in terms of mIoU/ave.F1. Additionally, the adoption of the hybrid loss function at the side and final output layers provided a further improvement of 0.2% in both mIoU and ave.F1. The inclusion of ARFE enhanced the mIoU/ave.F1 by 0.8%/0.5%, while FCSM provided the most significant improvement, increasing mIoU and ave.F1 by 1.2% and 0.7%, respectively. The combination of ARFE and FCSM achieved the highest mIoU score of 76.7% and the highest ave.F1 score of 86.6%. When replacing our proposed explicit supervision mechanism, i.e., FCSM, with the popular self-optimization mechanism, i.e., the unsupervised feature long-range correlation module (UFCM), in the self-attention operation, a decrease of 0.5% mIoU and 0.4% ave.F1 was observed.

These quantitative results demonstrate the effectiveness of the functional modules and supervision mechanisms in SLCNet. Table IV also shows that SSM and ARFE are very lightweight, while FCSM has slightly more parameters due to the introduction of linear projection layers (i.e., fully connected layers).

As for the long-range dependency modeling which we are mostly concerned with in this paper, we conducted more ablation experiments by replacing the proposed FCSM with other modeling strategies. Taking the model equipped with the SSM, ARFE, and the hybrid loss function training procedure as the new baseline, the Vision Transformer based module (VTM), the Swin transformer based module (STM), and the proposed FCSM were compared. VTM and STM were modified from the popular ViT block [18] and Swin transformer block [19] to adapt to the proposed network, which takes the features generated from the backbone as input and outputs feature maps of the same resolution. In particular, we stacked two ViT blocks and Swin transformer blocks in VTM and STM, respectively, corresponding to the two correlation operations of horizontal and vertical directions in the proposed FCSM. As with the implementation of FCSM in Fig. 2, an additional skip connection and a convolutional block were also employed in both VTM and STM, followed with further addition and convolution operations on the output correlation features.

As shown in Table V, all the long-range modeling strategies effectively improve the segmentation results, except for VTM, which conducts a low-resolution (16× downsampled) correlation between different tokens, as mentioned above. In the proposed FCSM and STM, we modified the correlation unit to keep the same resolution as the downsampled feature maps. Compared with the unsupervised feature long-range correlation module (UFCM) and STM, the proposed FCSM obtains a greater improvement in mIoU and ave.F1, demonstrating that the explicitly supervised feature correlation is more effective and suitable for remote sensing segmentation tasks. Meanwhile, UFCM performs slightly better than the unsupervised STM in terms of mIoU and ave.F1, but has fewer parameters, which further demonstrates the advantage of the proposed design.

*E. Discussion*

*1) Pretrained Weights and Model Performance:* It is widely recognized that the performance of a deep learning model can be boosted by pretraining on a proper large-scale dataset. In remote sensing land-cover classification, large-scale open-source close-range image classification datasets, especially ImageNet [42], are often used to pretrain a land-cover classification model. The effectiveness of this strategy is clearly demonstrated in Table VI, where the models pretrained with ImageNet obtain significant accuracy improvements. However, would an available remote sensing dataset, such as one of the datasets used in our experiments, perform better as a pretraining dataset? Theoretically, a remote sensing dataset would be closer to the remote sensing target dataset. We therefore compared the performances when initializing the model with pretrained weights from different datasets. The results are listed in Table VII, where the target dataset is the Vaihingen dataset, and the pretraining datasets are the ImageNet dataset, Potsdam dataset, and DeepGlobe dataset. It can be observed that the pretraining strategy with the different datasets brings improvements in terms of mIoU/ave.F1, but the models pretrained on the two remote sensing datasets show only a slight advantage over the model without pretraining. Although the Potsdam dataset has the same semantic categories as the Vaihingen dataset, it is apparent that the results from the ImageNet pretrained weights are significantly better than the others. According to Fig. 11, where the IoU score of each category is shown, it can be seen that ImageNet improves the category of car the most. This is because there are more car samples in the ImageNet dataset, although with close-range view angles. The experimental results indicate that the quantity of the dataset (ImageNet has 10 million images) and the variety of samples (there are 1000 categories in ImageNet) may be the critical factors that bring significant segmentation accuracy gains.

*2) Efficiency:* We compared the efficiency of all the comparison methods and the proposed SLCNet on the test set of the Vaihingen dataset. The running times are reported in Table VIII. The speeds of all the methods are tolerable, considering that there are about $2.32 \times e^7$ pixels in the Vaihingen test set. The long-range correlation modeling boosted methods, i.e., Swin-UNet [20], Trans-UNet [21], ST-UNet [13], and the proposed SLCNet, among which ST-UNet is the slowest, require more time than the classic methods (among which DeepLabv3+ [40] is the fastest), due to the additional long-term dependency modeling time. Although the proposed SLCNet has the second-largest number of parameters, it is the second-most efficient method among the long-range correlation modeling boosted methods, and only Swin-UNet is faster. Currently, SLCNet is based on the relatively heavy U-Net backbone. If, in our future work, we use a lighter backbone, as in Swin-UNet, the complexity and efficiency should be further optimized.

## V. Conclusion

In this paper, a novel supervised long-range correlation method has been proposed for the first time, providing specific supervision information for long-range correlation modeling in remote sensing land-cover classification. The proposed SLCNet method was shown to be more effective than the other widely used unsupervised transformer/attention-based strategies. In addition, an adaptive receptive field feature extraction (ARFE) module is designed to capture the size-adaptive and detailed object features from multi-scale images, and a hybrid loss function is employed to optimize the segmentation results from both local and global perspectives during the training procedure. The effectiveness of each module/strategy and the full SLCNet method was demonstrated through a comparison with the state-of-the-art methods from both the remote sensing, medicine, and computer vision communities on three public remote sensing datasets. We hope that the proposed method, and especially the supervised long-range dependency modeling, will be a beneficial supplement to the related studies using the unsupervised long-term correlation approach.

## References

[1] C. Zhang et al., "Joint deep learning for land cover and land use classification," *Remote Sens. Environ.*, vol. 221, pp. 173–187, Feb. 2019.

[2] D. Marcos, D. Volpi, B. Kellenberger, and D. Tuia, "Land cover mapping at very high resolution with rotation equivariant CNNs: Towards small yet accurate models," *ISPRS J. Photogramm. Remote Sens.*, vol. 145, pp. 96–107, Nov. 2018.

[3] X. Huang, D. Wen, J. Li, and R. Qin, "Multi-level monitoring of subtle urban changes for the megacities of China using high-resolution multi-view satellite imagery," *Remote Sens. Environ.*, vol. 196, pp. 56–75, Jul. 2017

[4] ] M. Volpi and D. Tuia, "Deep multi-task learning for a geographically regularized semantic segmentation of aerial images," *ISPRS J. Photogramm. Remote Sens.*, vol. 144, pp. 48–60, Oct. 2018.

[5] A. Krizhevsky, I. Sutskever, and G. E. Hinton, "ImageNet classification with deep convolutional neural networks," in *Proc. Adv. Neural Inf. Process. Syst.*, 2012, pp. 1097–1105.

[6] D. Yu, and S. Ji, "A new spatial-oriented object detection framework for remote sensing images," *IEEE Trans. Geosci. Remote. Sens.*, vol. 60, pp. 1-16, 2021.

[7] J. Long, E. Shelhamer, and T. Darrell, "Fully Convolutional Networks for Semantic Segmentation," *IEEE Trans. Pattern Anal. Mach. Intell.*, vol. 39, no. 4, pp. 640-651, 2015.

[8] L.-C. Chen, G. Papandreou, I. Kokkinos et al., "DeepLab: Semantic image segmentation with deep convolutional nets, atrous convolution, and fully connected CRFs," *IEEE Trans. Pattern Anal. Mach. Intell.*, vol. 40, no. 4, pp. 834-848, 2017.

[9] F. I. Diakogiannis, F. Waldner, P. Caccetta et al., "ResUNet-a: A deep learning framework for semantic segmentation of remotely sensed data," *ISPRS J. Photogramm. Remote Sens.*, vol. 162, pp. 94-114, 2020.

[10] L. Mou, Y. Hua, and X. X. Zhu, "Relation matters: Relational context-aware fully convolutional network for semantic segmentation of high resolution aerial images," *IEEE Trans. Geosci. Remote. Sens.*, vol. 58, no. 11, pp. 7557–7569, 2020

[11] R. Niu, X. Sun, Y. Tian et al., "Hybrid multiple attention network for semantic segmentation in aerial images," *IEEE Trans. Geosci. Remote. Sens.*, vol. 60, pp. 1-18, 2021.

[12] C. Zhang, W. Jiang, Y. Zhang et al., "Transformer and CNN Hybrid Deep Neural Network for Semantic Segmentation of Very-High-Resolution Remote Sensing Imagery," *IEEE Trans. Geosci. Remote. Sens.*, vol. 60, pp. 1-20, 2022.

[13] X. He, Y. Zhou, J. Zhao et al., "Swin Transformer Embedding UNet for Remote Sensing Image Semantic Segmentation," *IEEE Trans. Geosci. Remote. Sens.*, vol. 60, pp. 1-15, 2022.

[14] Z. Li, X. Chen, J. Jiang et al., "Cascaded multiscale structure with self-smoothing atrous convolution for semantic segmentation," *IEEE Trans. Geosci. Remote. Sens.*, vol. 60, pp. 1-13, 2021.

[15] S. Ji, S. Wei, and M. Lu, "A scale robust convolutional neural network for automatic building extraction from aerial and satellite imagery," *Int. J. Remote Sens.*, vol. 40, no. 9, pp. 3308-3322, 2019.

[16] G. Deng, Z. Wu, C. Wang et al., "CCANet: Class-constraint coarse-to-fine attentional deep network for subdecimeter aerial image semantic segmentation," *IEEE Trans. Geosci. Remote. Sens.*, vol. 60, pp. 1-20, 2021.

[17] J. Fu, J. Liu, H. Tian et al., "Dual Attention Network for Scene Segmentation," in *Proc. IEEE Int. Conf. Comput. Vis. Pattern Recog.*, 2019, pp. 3146-3154.

[18] A. Dosovitskiy, L. Beyer, A. Kolesnikov, D. Weissenborn, X. Zhai, T. Unterthiner, M. Dehghani, M. Minderer, G. Heigold, S. Gelly, J. Uszkoreit, and N. Houlsby, "An image is worth 16x16 words: Transformers for image recognition at scale," in *9th International Conference on Learning Representations, ICLR 2021, Virtual Event, Austria, May 3-7, 2021*. OpenReview.net, 2021.

[19] Z. Liu, Y. Lin, Y. Cao et al., "Swin Transformer: Hierarchical vision transformer using shifted windows." in *Proc. IEEE Int. Conf. Comput. Vision.*, 2021, pp. 10012-10022.

[20] H. Cao, Y. Wang, J. Chen, D. Jiang, X. Zhang, Q. Tian, and M. Wang, "Swin-UNet: UNet-like pure transformer for medical image segmentation," Unpublished paper, 2021. [Online]. Available: https://arxiv.org/abs/2105.05537.

[21] J. Chen, Y. Lu, Q. Yu, X. Luo, E. Adeli, Y. Wang, L. Lu, A. L. Yuille, and Y. Zhou, "TransUNet: Transformers make strong encoders for medical image segmentation," Unpublished paper, 2021. [Online]. Available: https://arxiv.org/abs/2102.04306.

[22] Y. Zhang, H. Liu, and Q. Hu, "TransFuse: Fusing transformers and CNNs for medical image segmentation." Unpublished paper, 2021. [Online]. Available: https://arxiv.org/abs/2102.08005.

[23] T. Xiao, Y. Liu, Y. Huang et al., "Enhancing Multiscale Representations with Transformer for Remote Sensing Image Semantic Segmentation," *IEEE Trans. Geosci. Remote. Sens.*, vol. 61, pp. 1-16, 2023.

[24] L.-C. Chen, G. Papandreou, F. Schroff, and H. Adam, "Rethinking atrous convolution for semantic image segmentation," Unpublished paper, 2017. [Online]. Available: http://arxiv.org/abs/1706.05587

[25] Q. Zhao, J. Liu, Y. Li et al., "Semantic segmentation with attention mechanism for remote sensing images," *IEEE Trans. Geosci. Remote. Sens.*, vol. 60, pp. 1-13, 2021.

[26] R. Guan, M. Wang, L. Bruzzone et al., "Lightweight Attention Network for Very High Resolution Image Semantic Segmentation," *IEEE Trans. Geosci. Remote. Sens.*, 2023.

[27] S. Ji, S. Wei, and M. Lu, "Fully convolutional networks for multisource building extraction from an open aerial and satellite imagery data set," *IEEE Trans. Geosci. Remote. Sens.*, vol. 57, no. 1, pp. 574-586, 2018.

[28] L. Ding, J. Zhang, and L. Bruzzone, "Semantic segmentation of large-size VHR remote sensing images using a two-stage multiscale training architecture," *IEEE Trans. Geosci. Remote. Sens.*, vol. 58, no. 8, pp. 5367-5376, 2020.

[29] W. Chen, Z. Jiang, Z. Wang et al., "Collaborative global-local networks for memory-efficient segmentation of ultra-high resolution images." in *Proc. IEEE Int. Conf. Comput. Vis. Pattern Recog.*, 2019, pp. 8924-8933.

[30] A. Vaswani et al., "Attention is all you need," in *Proc. Adv. Neural Inf. Process. Syst.*, 2017, pp. 5998–6008.

[31] K. He, X. Zhang, S. Ren, and J. Sun, "Deep residual learning for image recognition," in *Proc. IEEE Int. Conf. Comput. Vis. Pattern Recog.*, 2016, pp. 770-778.

[32] X. Yang, J. Yan, W. Liao et al., "SCRDet++: Detecting small, cluttered and rotated objects via instance-level feature denoising and rotation loss smoothing," *IEEE Trans. Pattern Anal. Mach. Intell.*, 2022.

[33] M. Berman, A. R. Triki, and M. B. Blaschko, "The Lovász-softmax loss: A tractable surrogate for the optimization of the intersection-over-union measure in neural networks." In *Proc. IEEE Int. Conf. Comput. Vis. Pattern Recog.*, 2018, pp. 4413-4421.

[34] ISPRS 2D Semantic Labeling Dataset, https://www.isprs.org/education/benchmarks/UrbanSemLab/.

[35] Y. Liu, D. M. Nguyen, N. Deligiannis, W. Ding, and A. Munteanu, "Hourglass-shape network based semantic segmentation for high resolution aerial imagery," *Remote. Sens.*, vol. 9, no. 6, p. 522, 2017.

[36] I. Demir, K. Koperski, D. Lindenbaum et al., "DeepGlobe 2018: A challenge to parse the earth through satellite images." in *Proc. IEEE Int. Conf. Comput. Vis. Pattern Recog.,* 2018, pp. 172-181.

[37] R. Zuo, G. Zhang, R. Zhang et al., "A Deformable Attention Network for High-Resolution Remote Sensing Images Semantic Segmentation," *IEEE Trans. Geosci. Remote. Sens.*, vol. 60, pp. 1-14, 2021.

[38] H. Zhao, J. Shi, X. Qi et al., "Pyramid Scene Parsing Network," in *Proc. IEEE Int. Conf. Comput. Vis. Pattern Recog.*, 2017, pp. 2881-2890.

[39] G. Lin, A. Milan, C. Shen et al., "RefineNet: Multi-path Refinement Networks for High-Resolution Semantic Segmentation," in *Proc. IEEE Int. Conf. Comput. Vis. Pattern Recog.*, 2017, pp. 1925-1934.

[40] L. C. Chen, Y. Zhu, G. Papandreou et al., "Encoder-Decoder with Atrous Separable Convolution for Semantic Image Segmentation," in *Proc. Eur. Conf. Comput. Vis.*, 2018, pp. 801-818.

[41] O. Ronneberger, P. Fischer, and T. Brox, "U-Net: Convolutional networks for biomedical image segmentation," in *Proc. Int. Conf. Med. Image Comput. Comput.-Assist. Intervent.*, 2015, pp. 234–241.

[42] A. Krizhevsky, I. Sutskever, and G. E. Hinton, "ImageNet classification with deep convolutional neural networks," *Adv. Neural Inf. Process. Syst.*, vol. 25, pp. 1097-1105, 2012.